%% file: VOIN_ICCV.tex
\documentclass[10pt,twocolumn,letterpaper]{article}

\usepackage{iccv}
\usepackage{times}
\usepackage{epsfig}
\usepackage{graphicx}
\usepackage{amsmath}
\usepackage{amssymb}
\usepackage{comment}
\usepackage{multirow}
\usepackage{booktabs}
\usepackage{amsmath,amssymb}
\usepackage{graphicx}
\usepackage{subfigure}
\usepackage{wrapfig}
\usepackage{color}

\newcommand{\bX}[0] {{\bf X}}
\newcommand{\bY}[0] {{\bf Y}}
\newcommand{\bM}[0] {{\bf M}}

\usepackage[pagebackref=true,breaklinks=true,letterpaper=true,colorlinks,bookmarks=false]{hyperref}

\iccvfinalcopy 

\def\iccvPaperID{4194} 

\ificcvfinal\fi
\begin{document}
\bibliographystyle{unsrt}
	
\title{Occlusion-Aware Video Object Inpainting}

\author{
 Lei Ke$^1$\hspace{1.0cm}Yu-Wing Tai$^{2}$\hspace{1.0cm}Chi-Keung Tang$^1$
 \vspace{0.1cm}\\
 $^1$The Hong Kong University of Science and Technology\hspace{1.0cm}$^2$Kuaishou Technology
 \\
\texttt{\footnotesize \{lkeab,cktang\}@cse.ust.hk, yuwing@gmail.com}} 

\maketitle
\ificcvfinal\thispagestyle{empty}\fi
	\begin{abstract}	
		\input{abstract.tex}
	\end{abstract}
	
	\vspace{-2mm}
   \section{Introduction}
   \input{introduction.tex}
   
   \section{Related Work}
   \input{related_work.tex}

   \section{Video Object Inpainting Network}
   \input{network.tex}
   
	
	\section{Experiments}
	\input{experiments.tex}
	
	
	\section{Conclusion}
	\input{conclusion.tex}

	{\small
		\bibliography{bib}
	}
	
\end{document}

%% file: abstract.tex
Conventional video inpainting is neither object-oriented nor
occlusion-aware, making it liable to obvious artifacts when large occluded object regions are inpainted. This paper presents occlusion-aware video object inpainting, which recovers both the complete shape and appearance for occluded objects in videos given their visible mask segmentation.

To facilitate this new research, we construct the first large-scale video object inpainting benchmark {\em YouTube-VOI} to provide realistic occlusion scenarios with both occluded and visible object masks available. Our technical contribution VOIN jointly performs video object shape completion and occluded texture generation. In particular, the shape completion module models long-range object coherence while the flow completion module  recovers accurate flow with sharp motion boundary, for propagating temporally-consistent texture to the same moving object across frames. For more realistic results, VOIN is optimized using both T-PatchGAN and a new spatio-temporal attention-based multi-class discriminator.

Finally, we compare VOIN and strong baselines on YouTube-VOI. Experimental results clearly demonstrate the efficacy of our method including inpainting complex and dynamic objects. VOIN degrades gracefully with inaccurate input visible mask. 

%% file: introduction.tex
\footnotetext[1]{Project page is at \url{https://lkeab.github.io/voin}.}
\footnotetext[2]{This research is supported in part by the Research Grant Council of the Hong Kong SAR under grant no. 16201420 and Kuaishou Technology.}

\begin{figure*}[!t]
	\centering
	\vspace{-0.2in}
	\includegraphics[width=0.98\linewidth]{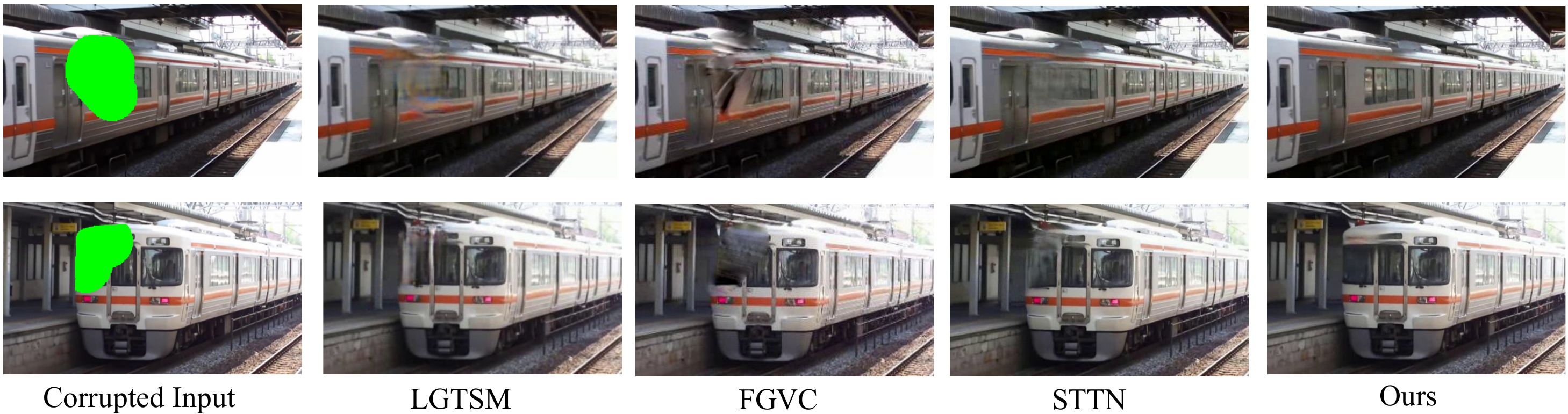}
	\caption{Video object inpainting results comparison with  state-of-the-art LGTSM~\cite{chang2019learnable}, FGVC~\cite{gao2020flow} and STTN~\cite{yan2020sttn}. Our VOIN takes corrupted video with free-form occlusion masks as input, and faithfully recovers the occluded object region while preserving spatial detail and temporal coherence. 
	}
	\label{fig:teaser_compare}
	\vspace{-0.2in}
\end{figure*}

Conventional video inpainting infers missing pixel regions by distilling information from remaining unmasked video regions. However, as shown in Figure~\ref{fig:teaser_compare}, these models~\cite{chang2019free,gao2020flow,yan2020sttn,xu2019deep} often fail to recover moving objects with large occlusion by wrongly inpainting the occluded region with irrelevant background colors and produce obvious artifacts. This is due to their lack of object and occlusion awareness.
On the contrary, our human visual system possesses powerful amodal perception ability to reason the complete structure of moving objects under occlusion, including the appearance of the invisible regions in high fidelity~\cite{aguiar2002developments,gae1979}.

To overcome the above limitations, we make the first significant attempt on occlusion-aware video object inpainting, which completes occluded video objects by recovering their shape and appearance in motion. While there exist object completion models, they are applicable only to single images~\cite{zhan2020self} in highly limited scenarios such as car and indoor furniture~\cite{ehsani2018segan,yan2019visualizing}. These single-image models do not utilize temporal coherence during mask and content generation, thus leading to temporal artifacts and unsmooth transition when directly applied to videos.

Training models for occlusion reasoning requires a large number and variety of occluded video objects with amodal mask annotations. One difficulty arises from the existing amodal datasets, which  come mostly in single images and are small~\cite{zhu2017semantic,follmann2019learning}, or cover very limited object classes~\cite{qi2019amodal}. Inspired by~\cite{zhan2020self}, our approach is trained in a self-supervised manner with only modal annotations.
To create realistic data in large quantity, we contribute the first large-scale video object inpainting benchmark with diverse occlusion patterns and object classes for both training and evaluation. Specifically,
we generate occlusion masks for video objects by high-fidelity simulation of overlapping objects in motion, thus taking into consideration object-like occlusion patterns, motion and deformation under various degrees of occlusion.
Our new \textbf{YouTube-VOI} dataset based on YouTube-VOS~\cite{xu2018youtube} contains 5,305 videos, a 65-category label set including common objects such as people, animals and vehicles, with over 2 million occluded and visible masks for moving video objects.

To infer invisible occluded object regions, we  propose the~\textbf{VOIN} (\textbf{V}ideo \textbf{O}bject \textbf{I}npainting \textbf{N}etwork), a unified multi-task framework for joint video object mask completion and object appearance recovery.
Our object shape completion module learns to infer complete object shapes from only visible mask regions and object semantics, while our appearance recovery module inpaints occluded object regions with plausible content.
To obtain pixel-level temporal coherency, we design a novel occlusion-aware flow completion module to capture moving video object and propagate consistent video content across even temporally distant frames, by enforcing flow consistency during the invisible object region generation. 
Furthermore, we augment the temporal patch-based GAN training process using a new multi-class discriminator with specially designed spatio-temporal attention module (STAM), which effectively accelerates model convergence and further improves inpainting quality.

Finally, we evaluate VOIN and strong adapted baselines on~\textit{YouTube-VOI} benchmark, where quantitative and qualitative results clearly demonstrate VOIN's advantages. This paper complements conventional video inpainting and facilitates future development of new algorithms on repairing occluded video objects.


\vspace{-0.1in}

%% file: related_work.tex
\paragraph{Video Inpainting.}
Previous works~\cite{wexler2007space,newson2014video,granados2012not,huang2016temporally,matsushita2006full} on  video inpainting fill arbitrary missing regions with visually pleasing content by learning spatial and temporal coherence, with deep learning based approaches~\cite{wang2019video,xu2019deep,zhang2019internal,kim2019deep,lee2019copy,oh2019onion,yan2020sttn,chang2019free,chang2019learnable} becoming mainstream in recent years.
The first deep generative model applied in video inpainting~\cite{wang2019video} combines 3D and 2D convolutions to produce temporally consistent inpainting content. 
To achieve better temporal consistency, in ~\cite{xu2019deep,zhang2019internal,gao2020flow} optical flows are used to guide propagation of information across frames. In~\cite{kim2019deep}  a temporal memory module is used with recurrent feedback.
For modeling long-range dependencies, in~\cite{lee2019copy}  frame-wise attention is applied on frames aligned by global affine transformation, and in~\cite{oh2019onion} the authors adopt pixel-wise attention to progressively fill the hole from its boundary.  
Temporal PatchGAN~\cite{chang2019free} based on SN-PatchGAN~\cite{yu2019free} and temporal shift modules~\cite{chang2019learnable} are proposed to further enhance inpainting quality.
Most recently, a spatio-temporal transformer is proposed in~\cite{yan2020sttn} for video completion by using a multi-scale patch-based attention module.

Although promising results have been achieved by the above methods, they still cannot satisfactorily recover the appearance of occluded video objects due to their lack of object and occlusion awareness.
On the other hand, our VOIN jointly optimizes the {\em amodal} shape and flow completion for occluded objects.
Different from previous flow completion~\cite{xu2019deep,zhang2019internal,gao2020flow} for random missing areas, our flow completion can faithfully recover the flow within occluded object regions. Using our large-scale synthetic data with accurate flow available for training, we can enforce and maintain strong flow consistency in spatial and temporal texture generation not previously possible in other methods. 
Together with the predicted amodal mask, we can thus produce more accurate object flow with sharper motion boundary than~\cite{xu2019deep,gao2020flow}.
Moreover, semantic information is incorporated into our proposed spatio-temporal multi-class discriminator, which makes the GAN training process faster and more stable, and further enhances the inpainting quality for  unseen regions.

\vspace{-10pt}

\paragraph{Amodal Object Completion.} 
In amodal object completion, visible masks of objects are given and the task is to complete the modal into amodal masks, which is different from amodal instance segmentation~\cite{li2016amodal,zhu2017semantic,hu2019sail,qi2019amodal}.
Previous amodal mask completion approaches make assumptions about the occluded regions, such as~Euler Spiral~\cite{kimia2003euler}, cubic Béziers~\cite{lin2016computational} and simple curves (straight lines and parabolas)~\cite{silberman2014contour}. These unsupervised methods cannot deal with objects with complex shapes.

Occlusion handling has also been extensively studied~\cite{sun2005symmetric,winn2006layout,gao2011segmentation,chen2015parsing,yang2011layered,hsiao2014occlusion,ke2020gsnet,kortylewski2020compositional,qi2021occluded}, especially in object detection~\cite{zhou2018bi,wang2018repulsion,kar2015amodal,wang2020robust}, segmentation~\cite{ke2021bcnet} and tracking~\cite{yang2005real,shu2012part,zhang2014partial,hua2014occlusion}, but most do not consider recovering the appearance of the occluded objects.
Among the prior amodal object completion works with appearance recovery, Ehsani~\textit{et al}~\cite{ehsani2018segan} generate the occluded parts of objects using Unet~\cite{ronneberger2015u} by leveraging about 5,000 synthetic images restricted to indoor scenes such as kitchen and living room.
Yan~\textit{et al}~\cite{yan2019visualizing} recover the appearance of occluded cars by synthesizing occluded vehicle dataset.
Zhan~\textit{et al}~\cite{zhan2020self} propose a self-supervised scene de-occlusion method PCNet which can complete the mask and content for  invisible parts of more common objects without amodal annotations as supervisions.
However, all of these methods are single image-based without considering temporal coherence and object motions. Extending them directly to  complex video sequence may easily lead to  unwanted temporal artifacts.
\vspace{-0.1in}

%% file: network.tex
\begin{figure*}[!t]
	\vspace{-0.3in}
	\centering
	\includegraphics[width=1.0\linewidth]{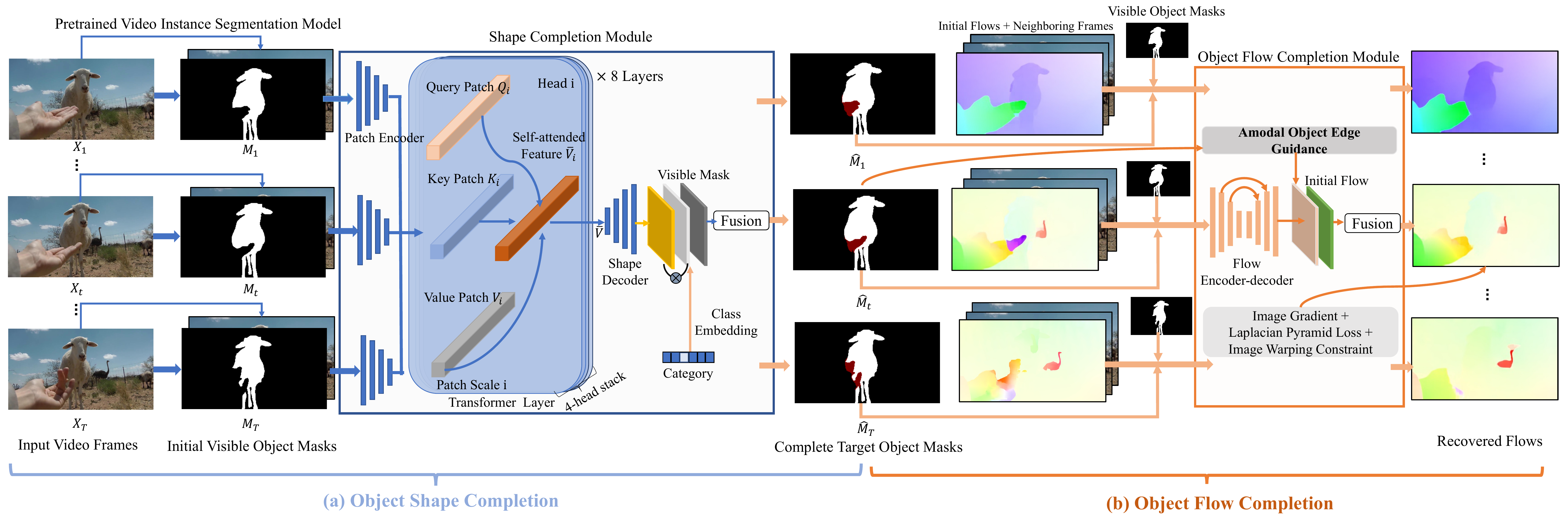}
	\vspace{-0.2in}
	\caption{\textbf{(a)} Object shape completion, which associates transformed temporal patches and object semantics;~\textbf{(b)} Object flow completion, which recovers complete object flow subject to the amodal object contours. Red indicates occluded region.}
	\label{fig:pipeline1}
	\vspace{-0.25in}
\end{figure*}

Given $\bX_{1}^{T}=\{X_1, X_2, ..., X_T\}$ as 
an input video sequence with frame length $T$, frame resolution $H \times W$, and $\bM_{1}^{T}=\{M_1, M_2, ..., M_T\}$ denotes 
the corresponding frame-wise binary masks for the visible regions of the target occluded object, we formulate the video object inpainting problem as self-supervised learning to infer complete object masks $\hat{\bM}_{1}^{T}=\{\hat{M}_1, \hat{M}_2, ..., \hat{M}_T\}$ and produce completed video frames $\bY_{1}^{T}=\{Y_1, Y_2, ..., Y_T\}$ with realistic amodal object content.

\begin{figure*}[!t]
	\centering
	\vspace{-0.2in}
	\includegraphics[width=1.0\linewidth]{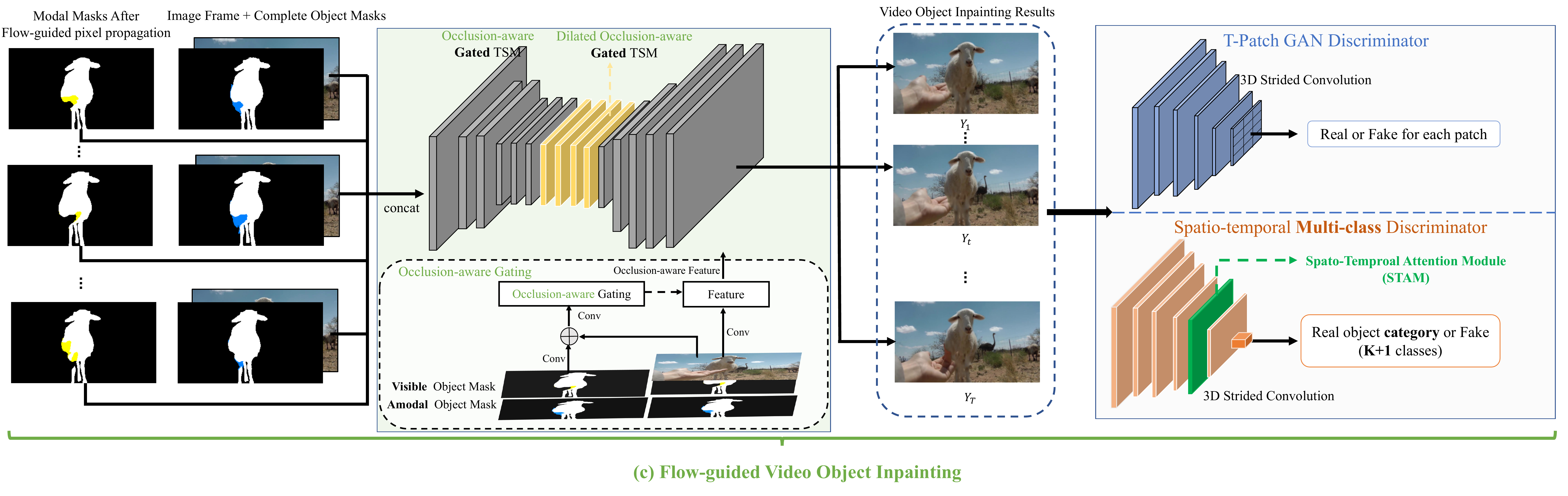}
	\vspace{-0.2in}
	\caption{
	\textbf{(c)}~Flow-guided video object inpainting with the occlusion-aware gating scheme.  Figure~\ref{fig:occlusion_aware} further illustrates the operation of our occlusion-aware temporal shift module (TSM). We adopt T-patch Discriminator~\cite{chang2019free,yu2019free} and propose a multi-class discriminator with spatio-temporal attention module (STAM) to regularize the GAN training. STAM is detailed in Figure~\ref{fig:STA}. Yellow marks the region filled by known pixels tracked by optical flow; blue regions are unseen/occluded regions to be inpainted in the final stage.}
	\label{fig:pipeline2}
	\vspace{-0.25in}
\end{figure*}

Figures~\ref{fig:pipeline1} and~\ref{fig:pipeline2} together depict the whole pipeline of our proposed video object inpainting approach VOIN, which consists of the following the three stages:
a)~{\em object shape completion}: we compute the amodal object shapes based on its visible object content (Section~\ref{shape_completion}); 
b)~{\em object flow completion}: complete object flow is  estimated with sharp motion boundary  under the guidance of amodal object contour (Section~\ref{flow_completion});  c)~{\em flow-guided video object inpainting}: with the completed object and flow within its contour,  motion trajectories are utilized to
warp pertinent pixels to 
inpaint the corrupted frames. 
To generate highly plausible video content, we improve temporal shift module 
by making it occlusion-aware and use multi-class discriminator with spatio-temporal attention,
instead of only using single-image completion techniques as in~\cite{gao2020flow,xu2019deep} (Section~\ref{sec:flow_guide}).

\vspace{-0.05in}
\subsection{Occlusion-Aware Shape Completion} 
\label{shape_completion}
Given an input video sequence, it is easy to obtain the modal masks for the target occluded object using the existing video object/instance segmentation~\cite{ventura2019rvos,robinson2020learning,yang2019video}. However, learning the full completion video masks of occluded instances is very difficult due to diverse object shapes and occlusion patterns. 

To address this problem, we propose a novel object shape completion module (see Figure~\ref{fig:pipeline1}(a)), which recovers amodal segmentation masks for the occluded video object in a self-supervised training scheme.
Our shape module with 8 transformer layers is inspired from the recent spatio-temporal transformers in video understanding~\cite{wang2020end,yu2020spatio,aksan2020attention,yan2020sttn,kim2018spatio} for capturing long-range spatio-temporal coherence. 
Each transformer layer has a multi-head structure to deal with the multi-scaled embeded image patches transformed from the whole input video sequence, followed by the scaled dot-product attention mechanism~\cite{vaswani2017attention}, which models temporal shape associations of the same occluded object among both the neighboring and distant encoded spatial feature patches in parallel.

Specifically, suppose there are $k$ heads in the transformer layer, then 
we compute self-attended feature ${\bar{V}}$:
\begin{equation}
\bar{V} =~\text{Multihead}(Q, K, V) = f_{c}([\bar{V}_{i}]_{i=1}^{k}),
\end{equation} 
\vspace{-0.2in}
\begin{equation}
\bar{V}_{i} = \text{SelfAtt}(Q_{i}, K_{i}, V_{i}) = \frac{\mathit{softmax}(Q_{i}K_{i}^T)}{\sqrt{d_k}}V_{i},
\end{equation} 
where~$\bar{V}_{i}$ is the self-attended feature on the $i$-th head, $Q_{i}$, $K_{i}$, $V_{i}$ are respectively the query, key and value embedding matrices for these spatial feature patches with total size $T\times~H/r_{1}\times~W/r_{2}$, frame resolution is $H \times W$, $r_{1}$ and $r_{2}$ are patch size, $d_k$ is the dimension for query patch features, and $f_{c}$ are convolutional layers merging the outputs from the~$k$ heads.  
$\bar{V}$ is then passed through the frame-level shape decoder for up-sampling, which is combined with the class embedding feature multiplied by the visible object masks for incorporating semantics and spatial shape prior. Finally, the merged features are refined by the fusion convolution layers to produce the amodal object shape masks.

\subsection{Occlusion-Aware Flow Completion} 
\label{flow_completion}
Our flow completion algorithm first computes the initial optical flows 
and then focuses on recovering the flow fields within the completed occluded object region subject to the amodal object contour. 

In~Figure~\ref{fig:pipeline1}(b), the flow generator adopts Unet~\cite{ronneberger2015u} encoder-decoder structure with skip connections from encoders to the corresponding layers in the decoder, which takes neighboring image frames, initial flow, visible and amodal object masks as inputs $x$.
Instead of computing the recovered flow directly, we formulate flow completion as a residual learning problem~\cite{he2016deep}, where $\phi (x) := O - \bar{O}$, $O$ is the desired flow output, $\bar{O}$ is the initial corrupted flow, and~$\phi (x)$ represents the flow residue learned by the encoder-decoder generator. This formulation effectively reduces the training difficulty for dense pixel regression.

To recover accurate object flow with sharp motion boundary, especially for the occluded region, we incorporate amodal object contour to guide the flow prediction process by enforcing flow smoothness within the complete object region, as flow fields are typically piecewise smooth, where gradients are small except along the distinct object motion boundaries. To effectively regularize the flow completion network, instead of simply adopting L1 regression loss between predictions and ground-truth flows as in~\cite{xu2019deep}, we additionally utilize the image gradient loss, Laplacian Pyramid loss~\cite{bojanowski2017optimizing} and image warping loss of hallucinated content for joint optimization, which further promote the precision of flow prediction and are detailed in section~\ref{optimization}.

\vspace{-0.1in}
\subsection{Flow-Guided Video Object Inpainting}
\label{sec:flow_guide}
The resulting completed object flow above is employed to build dense pixels correspondences across frames, which is 
essential since previously occluded regions in the current frame may be disoccluded and become visible in a distant frame, especially for objects in slow motion, which is very difficult for a generative model to handle such long-range temporal dependencies.

We follow~\cite{xu2019deep,gao2020flow} using forward-backward cycle consistency threshold (5 pixels) to filter out unreliable flow estimations and warp 
pixels bidirectionally to fill the missing regions based on the valid flow.
The main difference is that we only warp pixels within the foreground object region, which guarantees the occluded areas are not filled by any background colors while reducing the overall computational burden. Figure~\ref{fig:pipeline2} highlights the regions in yellow within a completed mask tracked by optical flow.

To fill in remaining pixels after the above propagation (i.e., the blue regions in Figure~\ref{fig:pipeline2}), which can be in large numbers for previously heavily occluded objects,
we propose to train an occlusion-aware gated generator to inpaint the occluded regions of videos objects, where the gating feature is learned under the guidance of both amodal object masks and occlusion masks, and two spatio-temporal discriminators with multi-class adversarial losses. As usual, the discriminators will be discarded during testing.

\vspace{-0.1in}
\label{inpainting_unseen}
\subsubsection{Occlusion-Aware TSM}
We adopt the residual Temporal Shift Module (TSM)~\cite{lin2019tsm,chang2019learnable} as our building blocks here, which shift partial channels along the temporal dimension to perform joint spatio-temporal feature learning, and achieve the performance of 3D convolution at 2D CNN's complexity. 

\begin{figure}[!h]
	\centering
	\vspace{-0.2in}
	\includegraphics[width=1.0\linewidth]{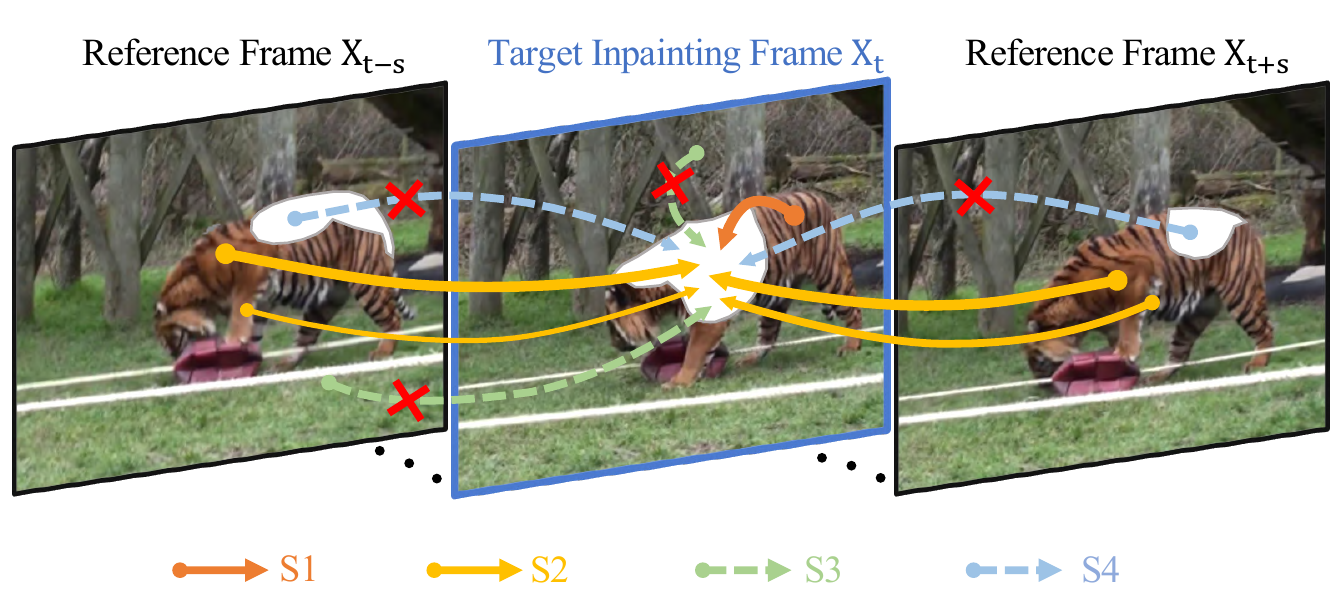}
	\caption{ Illustration of our occlusion-aware gating scheme. To inpaint the occluded region (in white) of the tiger in target frame $X_{t}$, 
		VOIN only learns from valid visible object parts along spatio-temporal dimension indicated by S1 and S2, while excluding irrelevant background pixels or occluded pixels by S3 and S4.}
	\label{fig:occlusion_aware}
	\vspace{-0.2in}
\end{figure}

However, the original TSM treats all feature points equally, making no difference between
visible and occluded regions belonging to the same object, and failing to distinguish invalid feature points from corrupted hole areas. Thus, to make TSM occlusion-aware and learn a dynamic feature selection mechanism for different spatial locations, we guide the gated feature learning process~\cite{yu2019free} with both amodal object masks and occlusion masks,
thus making our improved or occlusion-aware model capable of reasoning the occluded regions from other visible parts along the spatio-temporal dimension as illustrated in Figure~\ref{fig:occlusion_aware}. 

Specifically, the generator in Figure~\ref{fig:pipeline2} has the encoder-decoder structure with Occlusion-aware TSM replacing all vanilla convolutions layers, which  have a larger temporal receptive field~$n$ than original setting~\cite{lin2019tsm} and can be formulated as
\begin{equation}
\mathit{Gate}_{\text{occ}}^{x,y}(t)={\textstyle \sum}_{x,y}
W_g\cdot I_{t}^{x,y} +  \bar{f}_{t}^{x,y}(\hat{M}_{t}^{occ}, \hat{M}_{t}),
\end{equation}
\begin{equation}
\small
\mathcal{S}^{x,y}_t=
{\textstyle \sum}_{x,y}W_f\cdot \mathit{TSM}(I_{t-n}^{x,y},\cdots,I_{t}^{x,y},\cdots, I_{t+n}^{x,y}),
\end{equation}
\begin{equation}
\mathit{Out}^{x, y}_t= \sigma (\mathit{Gate}^{x,y}_{\text{occ}}(t)) \odot \phi(\mathcal{S}^{x,y}(t)),
\end{equation}
where $\mathit{Gate}_{occ}^{x,y}$ serves as a soft attention map (for identifying occluded/visible/background areas) on the feature volume $\mathcal{S}^{x,y}_t$ output by the TSM,~$\bar{f}$ are convolutional layers fusing the occlusion mask $\hat{M}_{t}^{occ}$ and complete object mask $\hat{M}_{t}$, $W_g$ and $W_f$ are respectively the kernel weights for gating convolution and shift module, and $I_{t}^{x,y}$ and $\mathit{Out}^{x, y}_t$ respectively denote the input and final output activation at $(t, x, y)$, $\sigma$ is sigmoid function, and $\phi$ is the ReLU function.

\vspace{-0.1in}
\subsubsection{Multi-Class Discriminator with STAM} 
To make object inpainting results more realistic, we adopt two discriminators to simultaneously regularize the GAN training process. The first discriminator considers video perceptual quality and temporal consistency, while the second considers object semantics based on both the global and local features since the occlusion holes may appear anywhere in the video with irregular shape.

\begin{figure}[!h]
	\centering
	\includegraphics[width=1.0\linewidth]{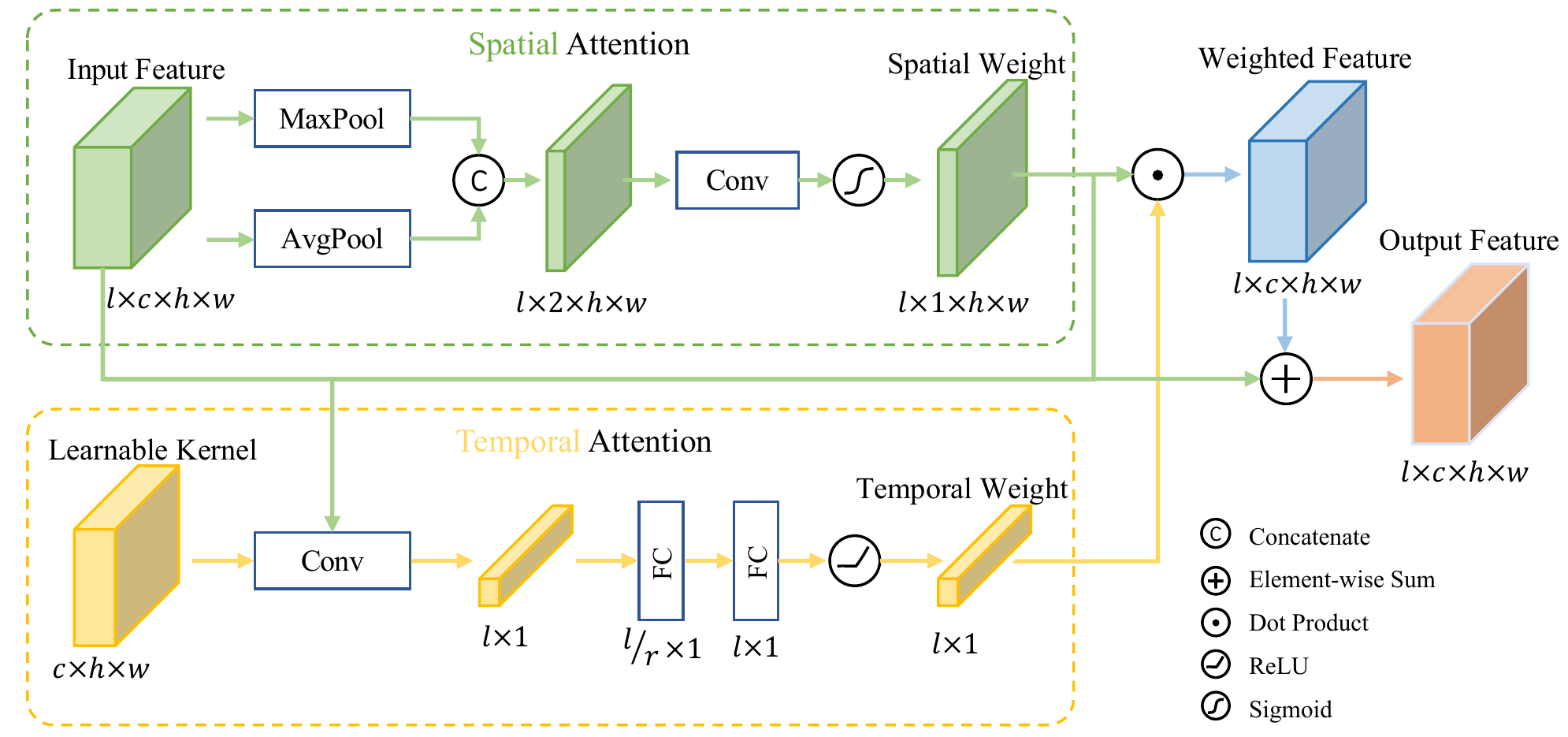}
	\caption{The design of spatio-temporal attention module (STAM), where two parallel branches are respectively used for computing spatial and temporal attention weights. The weighted feature forms a residual connection with the original input for final output.}
	\label{fig:STA}
	\vspace{-0.1in}
\end{figure}

We adopt T-PatchGAN as the first discriminator~\cite{chang2019free,yan2020sttn}. For the second discriminator, we propose a new spatio-temporal attention-based multi-class discriminator, which classifies the category of the inpainting object into one of the $K$  real classes and an additional fake class, by picking the most relevant frames from an input video while focusing on their discriminative spatial regions. Figure~\ref{fig:pipeline2} shows that the multi-class discriminator is composed of six 3D convolution layers (kernel size 3$\times$5$\times$5) with a spatio-temporal attention module (STAM) embedded above the $4$th layer. This STAM design is inspired by~\cite{hu2018squeeze,woo2018cbam}.  Figure~\ref{fig:STA} shows the parallel branches for spatial and temporal attention.

Although the deep layers of T-PatchGAN~\cite{chang2019free} can cover whole videos with its large receptive field, we find the  multi-class discriminator augmented with object semantics and STAM effectively accelerates model convergence and further improves inpainting quality (see Table~\ref{tab:voin_ab}).

\subsection{Optimization Objectives}
\label{optimization}
We train the multi-task VOIN framework in joint optimization with comprehensive objectives designed to produce sharp and spatio-temporal conherent video content, which are respectively shape completion loss $\mathcal{L}_\text{shape}$, flow completion loss $\mathcal{L}_\text{flow}$, and appearance recovery loss $\mathcal{L}_\text{app}$.

For video object shape completion, in addition to  the traditional binary cross entropy (BCE) loss, we also adopt dice loss~\cite{milletari2016v} to resolve the imbalance between the number of foreground
and background pixels, since the occluded regions sometimes only occupy a small area in the whole image. Thus $L_\text{shape}$ is formulated as
\begin{equation}
\mathcal{L}_{\text{shape}} = \mathcal{L}_{\text{BCE}}(M^{\prime},  \hat{M}) + \lambda_1\mathcal{L}_{\text{Dice}}(M^{\prime}_\text{occ}, \hat{M}_\text{occ}),
\label{eq:eq5}
\end{equation}
where $M^{\prime}$ and $M^{\prime}_\text{occ}$ respectively denote the predicted complete masks and the masks for occluded region, $\hat{M}$ and  $\hat{M}_\text{occ}$ are respectively the corresponding ground truth masks, and~$\lambda_1$ is the balance weight.

For dense object flow completion, we enforce the recovered flow with both the pixel-level accuracy and smooth flow field,  where $\mathcal{L}_\text{flow}$ is designed as
\begin{equation}
\mathcal{L}_{\text{flow}} = ||(1 + \hat{M})\odot(\hat{O}-O)||_{1} + \lambda_2\mathcal{L}_{\text{Lap}}(O, \hat{O}) + \mathcal{L}_{g} +\mathcal{L}_{w},
\end{equation}
\begin{equation}
\mathcal{L}_{g} =\hat{M}\odot(\lambda_3||(G_\Delta (O)-G_\Delta (\hat{O}))||_{1} +
\lambda_4||G_\Delta (O)||_{1}),
\end{equation}
where $\hat{O}$ and $O$ are respectively the predicted and ground-truth flow, $G_\Delta$ computes flow gradients in both horizontal and vertical directions, where we use~$\mathcal{L}_{g}$ to minimize the gradients for non-edge pixels inside the complete object to ensure smooth continuation while keeping object motion boundary with sharp transitions. $\mathcal{L}_\text{Lap}$ is used for preserving  details at different spatial scales~\cite{paris2011local}, and $\mathcal{L}_{w}$  supervises image warping consistency using the predicted flow.

To recover plausible object appearance, we optimize both the binary T-PatchGAN discriminator $D_{p}$ (to discriminate real or fake content)~\cite{chang2019free} and our multi-class global discriminator $D_\text{cls}$ (to classify category, with $K$  real classes and one fake class) using the spatio-temporal adversarial loss and semantic loss.
The optimization function $\mathcal{L}_\text{Dis}$ for the two discriminators is defined as 
\begin{equation}
\begin{footnotesize}
\begin{array}{l}
\mathcal{L}_\text{Dis} = \mathbb{E}_{x\sim p_{\text{data}}(x)}[1-D_{p}(x)]+ \mathbb{E}_{x\sim p_{\text{data}}(x)}[\log(D_{\text{cls}}(y|x))] \\ + \mathbb{E}_{z\sim p_{z}(z)}[1+D_{p}(G(z))] + \mathbb{E}_{z\sim P_{z}(z)}[\log(D_{\text{cls}}((K+1)|G(z)))],
\end{array}
\end{footnotesize}
\end{equation}
where $y \in \{1,...,K\}$, $ReLU$ is omitted for simplicity and the loss $~\mathcal{L}_\text{Gen}$ for inpainting generator is
\vspace{-0.1in}
\begin{equation}
\begin{footnotesize}
\begin{array}{l}
\mathcal{L}_\text{Gen} = -\mathbb{E}_{z\sim p_z(z)}[D_{p}(G(z))] -\mathbb{E}_{z\sim p_z(z)}[D_{cls}((K+1)|G(z))],
\end{array}
\end{footnotesize}
\end{equation}
The per-pixel content reconstruction loss $\mathcal{L}_\text{content}$ and appearance recovery loss $\mathcal{L}_\text{app} $ are respectively defined as
\begin{equation}
\mathcal{L}_\text{content} = ||\hat{M}\odot(Y^{\prime} - Y)||_{1} + \lambda_5 ||(1-\hat{M})\odot(Y^{\prime} - Y)||_{1},
\end{equation}
\begin{equation}
\begin{small}
\mathcal{L}_\text{app} = \mathcal{L}_\text{Dis} + \mathcal{L}_\text{Gen} + \lambda_6\mathcal{L}_\text{content},
\end{small}
\end{equation}
where $Y'$ and $Y$ are respectively the ground truth and predicted completed frame.
Thus, the overall optimization objectives are summarized as
\begin{equation}
\mathcal{L}_{\text{total}} = \mathcal{L}_{\text{shape}} + \lambda_\text{flow}\mathcal{L}_{\text{flow}} + \lambda_\text{app}\mathcal{L}_\text{app}.
\end{equation}

%% file: experiments.tex
\vspace{-0.02in}
\subsection{Dataset and Evaluation Metrics}

\vspace{-0.07in}
\paragraph{Occlusion Inpainting Setting}
Since this paper focuses on inpainting  occluded regions of video objects, we propose a new inpainting setting which is different from previous ones for undesired object removal or arbitrary mask region inpainting. 
The fill-up regions are restricted to occluded regions of the target object, which can be given by user or our object shape completion module using the visible object content.
This setting is in line with 
real-world applications such as
video scene de-occlusion (Figure~\ref{fig:deocc}).

\vspace{-0.2in}

\paragraph{YouTube-VOI Benchmark.}
To support training and evaluation of our new video object inpainting task, we use the YouTube-VOS~\cite{xu2018youtube} dataset as our video source to construct our large-scale YouTube-VOI benchmark, which contains 5,305 videos (4,774 for training and 531 for evaluation) with resolution higher than 640 $\times$ 480, a 65-category label set including common objects such as people, animals and vehicles, and over 2 million occluded and visible masks.
\vspace{-0.1in}
\begin{figure}[!h]
	\centering
	\includegraphics[width=0.95\linewidth]{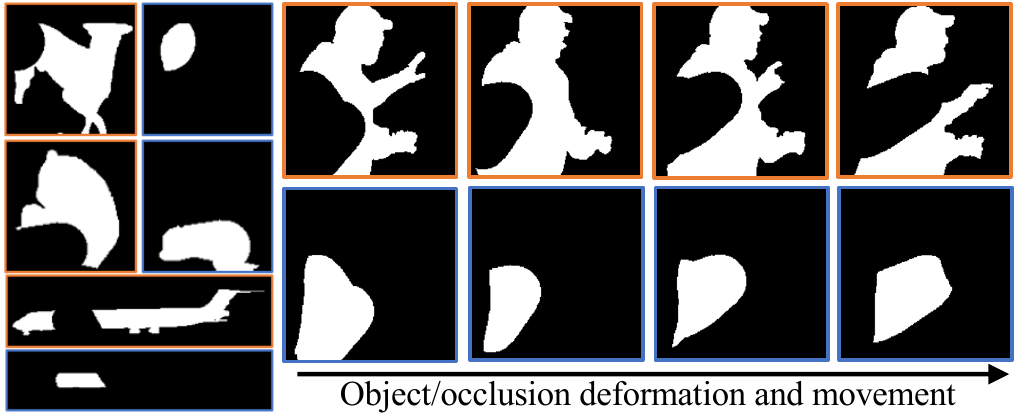}
	\caption{Sample visible masks (orange boxes) and occluded masks (blue boxes) for moving video objects generated by our algorithm with different object categories and occlusion patterns.}
	\label{fig:occ_mask}
	\vspace{-0.2in}
\end{figure}

We generate both the occlusion masks and visible masks for video objects by high-fidelity simulation of overlapping objects in motions (Figure~\ref{fig:occ_mask}), where we take into consideration various object-like occlusion patterns, occluder movements and shape deformations under various degrees of occlusion  from 10\% to 70\%.

Our YouTube-VOI  is a very challenging dataset for video object inpainting, which is representative of complex real-world scenarios in high diversity, including different realistic occlusions caused by vehicles, animals and human activities. Although amodal object masks are not annotated in Youtube-VOS, we show that the proposed VOIN model can still perform amodal object shape completion for video with only modal annotations, by incorporating the self-supervised training scheme in~\cite{zhan2020self} to learn object shape associations between frames, utilizing the annotated object semantics, and training on huge number of occlusions masks in various degrees and patterns of occlusion.

\vspace{-0.2in}

\paragraph{Evaluation Metrics.}
For evaluating video object inpainting quality, we use the widely adopted PSNR, SSIM and LPIPS metrics following~\cite{gao2020flow,xu2019deep}, where LPIPS~\cite{zhang2018unreasonable} is obtained using Alexnet~\cite{krizhevsky2012imagenet} as backbone with linear calibration on top of intermediate features as its default setting. 

\vspace{-10pt}

\paragraph{Implementation Details.}

We build our occlusion-aware inpainting generator adapted from the encoder-decoder structure in~\cite{chang2019learnable}. 
For data preprocessing, we resize video frames to 384 × 216 and randomly crop them to 320 × 180 with random horizontal flip.
We generate video object occlusion masks based on the free-from masks~\cite{yu2019free,chang2019free}.
For more network and implementation details, please refer to the supplementary materials.

\vspace{-0.05in}
\subsection{Comparison with State-of-the-arts}
Using the Youtube-VOI benchmark,
we compare VOIN with the most recent and relevant state-of-the-art video inpainting approaches and adapt their original input by additionally concatenating visible object masks: 1)~DFVI~\cite{xu2019deep}, which fills corrupted regions using pixel propagation based on the predicted complete flow; 2)~LGTSM~\cite{chang2019learnable}, where learnable shift module is designed for inpainting generator and T-PatchGAN discriminator~\cite{chang2019free} is utilized; 3)~FGVC~\cite{gao2020flow}, which uses a flow completion module guided by Canny edge extraction~\cite{canny1986computational} and connection~\cite{nazeri2019edgeconnect}; 4)~STTN~\cite{yan2020sttn}, which completes missing regions using multi-scale patch-based attention module. Note that both DFVI and FGVC conduct flow completion for inpainting, and fill the remaining unseen video regions using only image inpainting method~\cite{yu2018generative}.

\vspace{-0.2in}

\begin{figure}[!h]
	\centering
	\vspace{-0.2in}
	\includegraphics[width=0.92\linewidth]{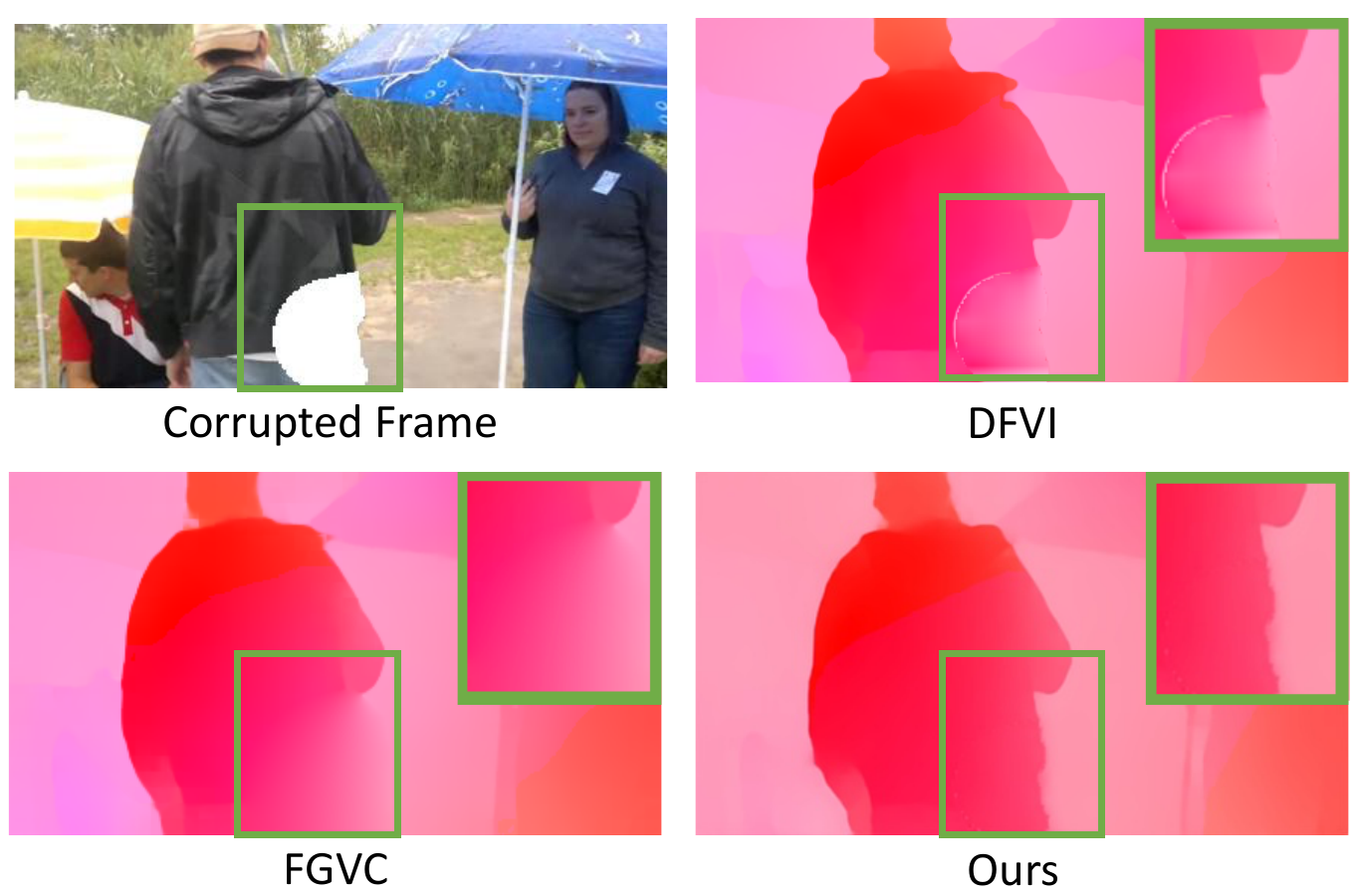}
	\caption{Flow completion results comparison with DFVI~\cite{xu2019deep} and FGVC~\cite{gao2020flow}. Our method completes object flow with sharper motion boundary and more natural piece-wise smooth transition with no visible seams within the hole compared to DFVI.
	}
	\label{fig:flow_compare_with}
	\vspace{-0.3in}
\end{figure}

\paragraph{Flow Completion Comparison.} 
We compare our object flow completion module with~\cite{xu2019deep,gao2020flow} both qualitatively and quantitatively. 
Table~\ref{tab:t1} shows the endpoint error (EPE) between the pseudo ground truth flow computed from the original, un-occluded videos using RAFT~\cite{teed2020raft} and the predicted completed flow on the missing regions. 
The quantitative result reveals that the proposed object flow completion module achieves significantly lower EPE errors than the previous flow completion networks in~\cite{xu2019deep,gao2020flow}.
Figure~\ref{fig:flow_compare_with} compares    flow completion results, where our proposed occlusion-aware flow completion module produces sharp motion boundary with a smooth flow field within the object contour, which shows the effectiveness of amodal object shape guidance and hybrid loss optimization. Although FGVC~\cite{gao2020flow} trained a separate flow edge connection network using EdgeConnect~\cite{nazeri2019edgeconnect}, their model still cannot complete large occlusion holes but generates blurred and ambiguous flow completion result. 

\begin{table}[!h]
	\vspace{-0.1in}
	\caption{Quantitative comparison on flow completion (EPE)  and inpainting quality (PSNR, SSIM and LPIPS) on Youtube-VOI benchmark with the state-of-the-art methods.}
	\vspace{0.05in}
	\centering
	\resizebox{1.0\linewidth}{!}{
		\begin{tabular}{ l | c | c | c | c | c}
			\toprule
			Model & Use flow? & EPE $\downarrow$ & PSNR $\uparrow$ & SSIM $\uparrow$ & LPIPS $\downarrow$\\
			\midrule
			DFVI~\cite{xu2019deep} & \checkmark & 4.79 & 44.91 & 0.952 & 0.099 \\
			LGTSM~\cite{chang2019learnable} & & - & 45.19 & 0.979 & 0.024 \\
			FGVC~\cite{gao2020flow}  & \checkmark & 3.69 & 43.90 & 0.924 & 0.065 \\
			STTN~\cite{yan2020sttn}  & & - & 45.97 & 0.986 & 0.020 \\
			\midrule
			Ours & & - & 46.33 &  0.989 & 0.013  \\
			Ours & \checkmark & \textbf{3.11} & \textbf{48.99} & \textbf{0.994} & \textbf{0.008}  \\
			\bottomrule
		\end{tabular}
	}
	\label{tab:t1}
	\vspace{-0.2in}
\end{table}

\vspace{-0.2in}

\paragraph{Quantitative  Results Comparison.}

Table~\ref{tab:t1} also reports quantitative comparison on inpainting quality under complex occlusion scenarios on the Youtube-VOI test set.
Compared to existing models, our VOIN substantially improves video reconstruction quality with both per-pixel and overall perceptual measurements, where our model outperforms the most recent STTN~\cite{yan2020sttn} and FGVC~\cite{gao2020flow} by a large margin, especially in terms of PSNR and LPIPS.
Our improved results show the effectiveness of our proposed occlusion-aware gating scheme and the multi-class discriminator with STAM.
On the other hand, the results produced by DFVI and FGVC are not on par with ours, especially for inpainting foreground object with large occlusion due to their incorrect flow completion and the lack of temporal consistency for generating video content (limited in using single image-based inpainting model DeepFill~\cite{yu2018generative}). 

\begin{figure*}[!t]
    \vspace{-0.25in}
	\centering
	\includegraphics[width=0.99\linewidth]{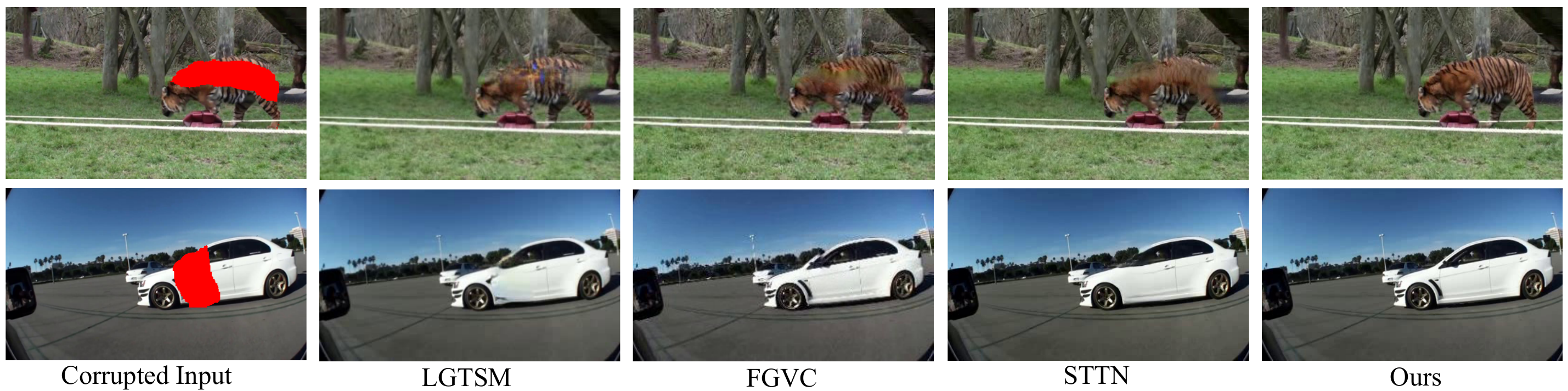} \\
	\caption{Qualitative comparison with  state-of-the-art video inpainting methods LGTSM~\cite{chang2019learnable}, FGVC~\cite{gao2020flow}, STTN~\cite{yan2020sttn} on Youtube-VOI. In particular, FGVC also adopts completed flow to guide the video inpainting process, but their results suffer from unnatural pixel transition due to the incorrect flow estimation. Zoom in for better view. Refer to the supplementary file for more qualitative comparison.}
	\label{fig:teaser_compare1}
	\vspace{-0.02in}
\end{figure*}

\begin{figure}[!h]
	\centering
	\includegraphics[width=1.0\linewidth]{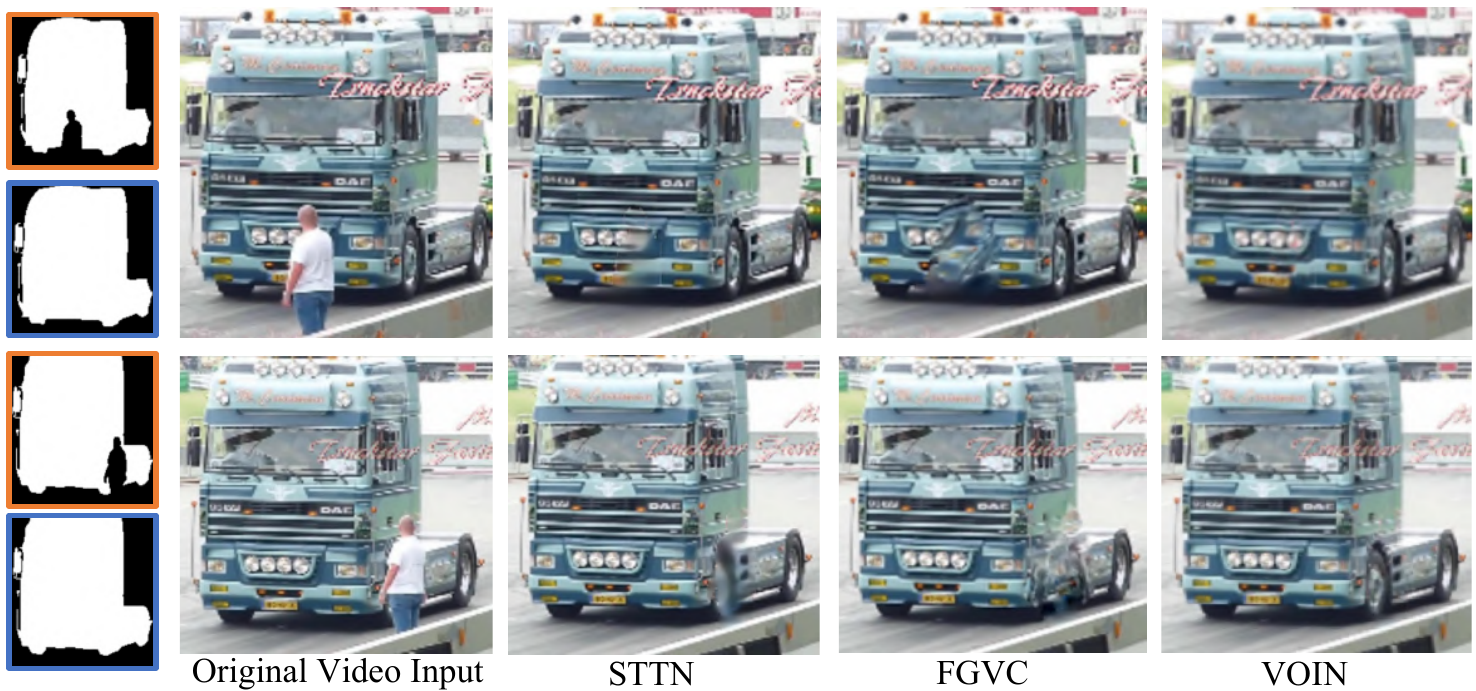}
	\vspace{-0.15in}
	\caption{Video scene de-occlusion results comparison, using the shape completion predictions by our VOIN. The left column contains both the visible masks (orange boxes) and predicted complete masks (blue boxes). VOIN detects and recovers the occluded region of the truck with faithful spatial details. The background leaking area (road and shadow under the truck) of the truck occluded by the person's lower body is filled by~\cite{yan2020sttn} for fair comparison. 
	}
	\label{fig:deocc}
	\vspace{-0.2in}
\end{figure}

\vspace{-0.2in}
\paragraph{Qualitative Results Comparison.} 
Figure~\ref{fig:teaser_compare1} shows sample video completion results for inpainting occluded video objects, where our occlusion-aware VOIN produces temporally coherent and visually plausible
content than previous methods~\cite{chang2019learnable,gao2020flow,yan2020sttn}.
Figure~\ref{fig:deocc} shows its application in video scene de-occlusion.
Refer to the supplementary video results for extensive qualitative comparison.


\begin{figure*}[!t]
	\vspace{-0.15in}
	\centering
	\includegraphics[width=0.99\linewidth]{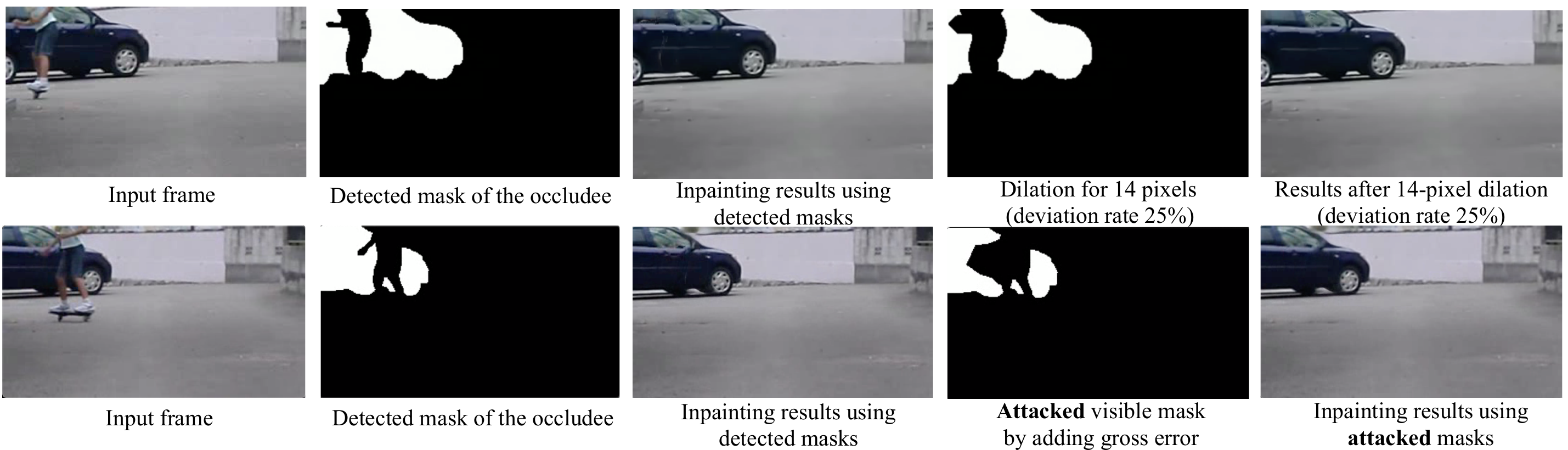} \\
	\vspace{-0.05in}
	\caption{Sample visual results of VOIN given inaccurate mask segmentation (dilation and gross segmentation errors), which show the robustness of VOIN. Full  results are available in the supplementary file.}
	\label{fig:teaser_compare2}
	\vspace{-0.25in}
\end{figure*}

\vspace{-5pt}
\subsection{Ablation Study}


\paragraph{Video Object Shape Completion.} 
To evaluate the effectiveness of different components in occlusion-aware object shape completion module, Table~\ref{tab:shape_comp} reports the corresponding ablation study by reducing transformer layers to 4 and removing semantic embedding and dice loss~\cite{milletari2016v}.
The spatial shape prior of object categories can greatly improve the mIoU by around 5\%, and the dice loss further promotes the completion performance on small occluded regions.
We also compare our video shape completion module with the image-based UNet trained on amodal annotation~\cite{zhan2020self}, where the large performance gap reveals the importance of learning object shape associations between frames.

\vspace{-0.2in}
\paragraph{Influence of Initial Mask Segmentation.}
Figure~\ref{fig:teaser_compare2} shows the influence of initial segmentation quality on the final object inpainting results in the presence of mask errors. Please refer to Section 4 of the supplemental file for full mask degradation experimental results and analysis.

\begin{table}[!t]
    \vspace{-0.2in}
	\caption{Ablation study on our \textit{object shape completion} module. B$_\text{S-n}$: video shape completion module with $n$ transformer layers and BCE loss. D: using dice loss. S: adding semantic guidance.}
	\vspace{0.05in}
	\centering
	\resizebox{0.85\linewidth}{!}{
		\begin{tabular}{ l | c | c | c | c | c}
			\toprule
			& Image-based & B$_\text{S-4}$ & B$_\text{S-8}$ & B$_\text{S-8}$ + D & B$_\text{S-8}$ + D +  S\\
			\midrule
			mIoU (\%) & 74.81 & 78.02 & 80.28 & 81.53 & \textbf{86.68} \\
			\bottomrule 
		\end{tabular}
	}
	\vspace{-0.2in}
	\label{tab:shape_comp}
\end{table}

\begin{figure}[!h]
	\centering
	\includegraphics[width=0.90\linewidth]{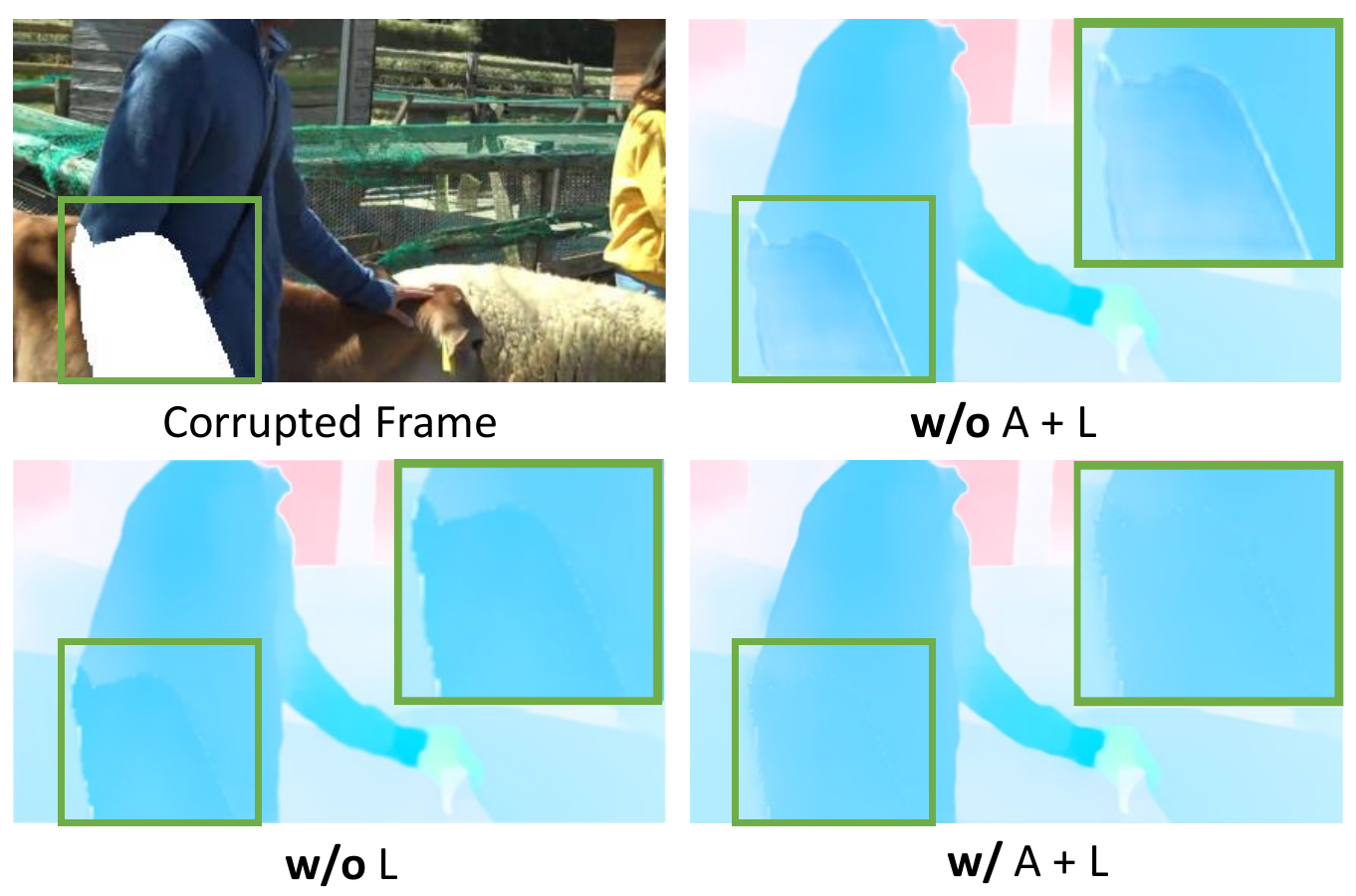}
	\vspace{-0.05in}
	\caption{Ablation results of our object flow completion module. A: The predicted amodal object shape guidance. L: Image gradient and warping loss. The corrupted region \textbf{w/o} A+L has a large portion of flow error, and the result \textbf{w/o} L has unnatural flow transition exhibiting visible seam within the completed foreground.}
	\label{fig:flow_compare}
	\vspace{-0.28in}
\end{figure}

\vspace{-15pt}
\paragraph{Video Object Flow Completion.} 
Table~\ref{tab:flow_comp} depicts the ablation experiment result on the proposed object flow completion module. We find that the end-to-end point flow error is reduced progressively by successively incorporating amodal shape guidance and hybrid loss constraints. Also, we show the flow completion comparison in Figure~\ref{fig:flow_compare} to visualize the effectiveness of each component. The amodal shape guidance prevents corruption due to irrelevant background flows, and the image gradient loss effectively retains the natural smooth flow transition within the completed object, although we observe that using gradient loss takes the flow completion network longer time to converge.

\vspace{-0.1in}
\begin{table}[!h]
	\caption{Ablation study on our \textit{object flow completion} module. B$_\text{F}$:  flow completion baseline using Unet structure for regressing flow directly. A: amodal object edge guidance. L: Gradient loss and image warping loss. }
	\vspace{0.03in}
	\centering
	\resizebox{0.85\linewidth}{!}{
		\begin{tabular}{ l | c | c | c}
			\toprule
			& B$_\text{F}$ & B$_\text{F}$ + A & B$_\text{F}$ + A + L  \\
			\midrule
			Flow completion (EPE) & 4.89 & 3.95 & \textbf{3.11} \\
			\bottomrule 
		\end{tabular}
	}
	\vspace{-0.3in}
	\label{tab:flow_comp}
\end{table}

\begin{table}[!h]
	\caption{Ablation study on different VOIN components. B$_\text{I}$:  VOIN baseline using the generator network~\cite{chang2019learnable}.  OG: our occlusion-aware gating scheme. TP: T-PatchGAN discriminator~\cite{chang2019free}. MD: multi-class discriminator. STAM: our spatio-temporal attention module. F: our object flow completion guidance.}
	\vspace{0.05in}
	\centering
	\resizebox{0.90\linewidth}{!}{
		\begin{tabular}{ l | c | c | c | c }
			\toprule
			Model & Use flow? & PSNR $\uparrow$ & SSIM $\uparrow$ & LPIPS $\downarrow$ \\
			\midrule
			B$_\text{I}$  & & 44.75 & 0.979 & 0.025 \\
			B$_\text{I}$  + OG &  & 45.25 & 0.982 & 0.019 \\
			B$_\text{I}$  + OG + TP & & 45.54 & 0.984 & 0.017 \\
			B$_\text{I}$  + OG + TP + MD & & 45.91 & 0.986 & 0.014 \\
			B$_\text{I}$  + OG + TP + MD + STAM & & 46.33 &  0.989 & 0.013 \\
			B$_\text{I}$ + OG + STAM  & \checkmark & 48.16 & 0.991 & 0.012 \\
			B$_\text{I}$ + OG + TP + MD + STAM  & \checkmark & \textbf{48.99} & \textbf{0.994} & \textbf{0.008} \\
			\bottomrule
		\end{tabular}
	}
	\vspace{-0.3in}
	\label{tab:voin_ab}
\end{table}

\subsubsection{VOIN Model}
\vspace{-0.05in}
To investigate how each component in VOIN contributes to the final video object inpainting performance, especially for the proposed occlusion-aware gating scheme and multi-class discriminator with spatio-temporal attention module (STAM), Table~\ref{tab:voin_ab} reports the ablation study results, and we analyze the effectiveness of our design choices as follows.

\vspace{-0.2in}
\paragraph{Effect of Occlusion-Aware Gating.} 
Table~\ref{tab:voin_ab} shows the importance of identifying occluded/visible/background areas on the input feature volume (illustrated in Figure~\ref{fig:occlusion_aware}), where our occlusion-aware gating scheme learns to infer the occluded object regions by attending to the visible spatio-temporal object regions without adversely affected by the background. This strategy significantly improves the perceptual similarity metric LPIPS by 24\% and improves the PSNR  from 44.75 to 45.25.

\vspace{-0.2in}

\paragraph{Effect of MD with STAM.} Multi-class discriminator (MD) incorporates semantics into  GAN training, which enables  more fine-grained object classification and thus produces more realistic video content, improving PSNR from 45.54 to 45.91. Furthermore, STAM enhances the overall performance by enabling the multi-class discriminator to focus on more discriminative feature region through learning spatio-temporal attention weights across video frames, which in turn improves the inpainting quality during the two-player adversarial process~\cite{goodfellow2014generative}.

\vspace{-0.18in}
\paragraph{Effect of Flow-Guided Pixel Propagation.}
The completed flow warps valid pixels to fill the missing regions in video frames, which greatly reduces the inpainting area and thus the difficulty. The last two rows of Table~\ref{tab:voin_ab} reflects the accuracy of the completed object flow with large performance gain, where the inpainting result of VOIN under flow guidance w/o adversarial training remains high quality. 
\vspace{-0.1in}

%% file: conclusion.tex
This paper proposes the new occlusion-aware video object inpainting task with the first large-scale video object inpainting benchmark {\em YouTube-VOI} for both training and evaluation. VOIN is a multi-task framework that completes shape and appearance for occluded objects in videos given their visible masks, which 
contains novel occlusion-aware shape and flow completion modules for propagating temporally-consistent object texture, and a spatio-temporal multi-class discriminator with STAM for enhancing object inpainting quality.
We compare VOIN with strong adapted baselines on~\textit{YouTube-VOI} benchmark and achieve competitive performance.
Our proposed VOIN may benefit many video applications such as video scene de-occlusion/manipulation, and improve video object tracking accuracy under heavy occlusion.